\documentclass[letterpaper]{article} 
\usepackage[]{aaai2026}  
\usepackage{times}  
\usepackage{helvet}  
\usepackage{courier}  
\usepackage[hyphens]{url}  
\usepackage{graphicx} 
\usepackage{natbib}  
\usepackage{caption} 
\usepackage{colortbl}
\usepackage{algorithm}
\usepackage{algorithmic}
\usepackage{amsmath} 
\usepackage{mathrsfs}  
\usepackage{mathtools} 
\usepackage{amsmath}  
\usepackage{amssymb}  
\usepackage{amsfonts} 
\usepackage{graphicx} 
\usepackage{booktabs} 
\usepackage{multirow}

\usepackage{newfloat}
\usepackage{listings}
\DeclareCaptionStyle{ruled}{labelfont=normalfont,labelsep=colon,strut=off} 
\floatstyle{ruled}
\newfloat{listing}{tb}{lst}{}
\floatname{listing}{Listing}
\title{ LiteUpdate: A  Lightweight Framework for Updating \\ AI-Generated Image Detectors}

\author{
    \small
    Jiajie Lu\textsuperscript{\rm 1},
    Zhenkan Fu\textsuperscript{\rm 1},
    Na Zhao\textsuperscript{\rm 1},
    Long Xing\textsuperscript{\rm 1,2},
    Kejiang Chen\textsuperscript{\rm 1}\equalcontrib,
    Weiming Zhang\textsuperscript{\rm 1}\equalcontrib,
    Nenghai Yu\textsuperscript{\rm 1}
}
\affiliations{
    \textsuperscript{\rm 1}University of Science and Technology of China, \textsuperscript{\rm 2}Shanghai AI Laboratory
}

\usepackage{bibentry}
\usepackage{pifont}         
\newcommand{\xmark}{\ding{55}} 

\begin{document}

\maketitle 

\begin{abstract}
The rapid progress of generative AI has led to the emergence of new generative models, while existing detection methods struggle to keep pace, resulting in significant degradation in the detection performance. This highlights the urgent need for continuously updating AI-generated image detectors to adapt to new generators.
To overcome low efficiency and catastrophic forgetting in detector updates, we propose LiteUpdate, a lightweight framework for updating AI-generated image detectors. 
LiteUpdate employs a representative sample selection module that leverages image confidence and gradient-based discriminative features to precisely select boundary samples. This approach improves learning and detection accuracy on new distributions with limited generated images, significantly enhancing detector update efficiency. Additionally, LiteUpdate incorporates a model merging module that fuses weights from multiple fine-tuning trajectories, including pre-trained, representative, and random updates. This balances the adaptability to new generators and  mitigates the catastrophic forgetting of prior knowledge. 
Experiments demonstrate that LiteUpdate substantially boosts detection performance in various detectors. Specifically, on AIDE, the average detection accuracy on Midjourney improved from 87.63\% to 93.03\%, a 6.16\% relative increase.

\end{abstract}

\section{Introduction}

Recently, the remarkable advances in generative models have given new impetus to the field of Generative AI. Generative models such as GANs \cite{goodfellow2014generative,karras2019style} and diffusion models \cite{ho2020denoising,song2020denoising,saharia2022photorealistic} have seen increasing use on public platforms, driven by their ability to generate high-quality images. However, these technologies also raise concerns due to their potential for misuse, including creating some fake news, manipulating public opinion, influencing political elections, and violating copyright laws \cite{epstein2023art}.

\begin{figure}[t]
    \centering
    \includegraphics[width=0.45\textwidth, height=0.35\textwidth]{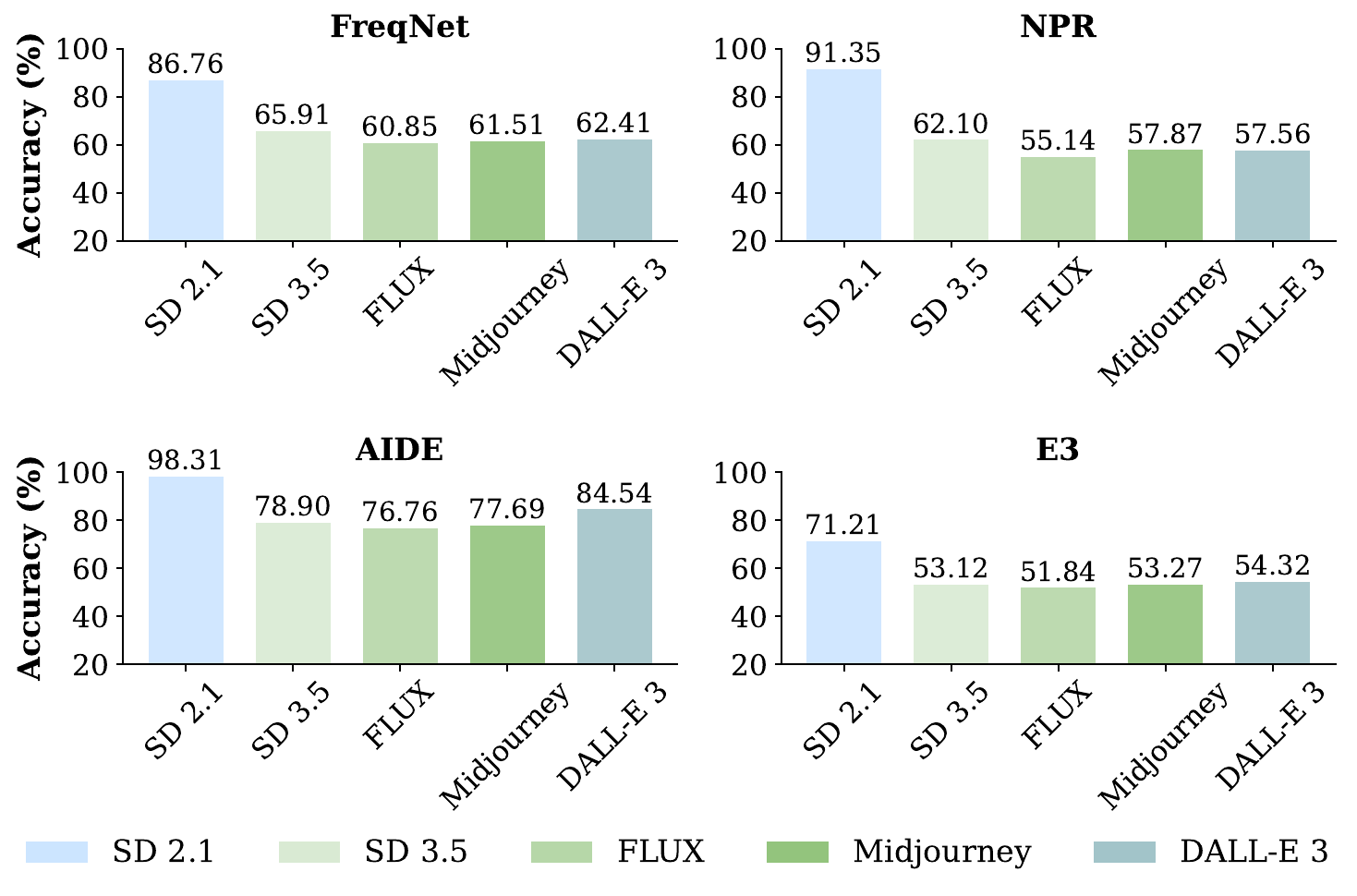}
    \caption{All traditional detectors pre-trained on SD 2.1 suffer a significant performance drop when applied to images from unseen generative models. Each subplot represents the detection performance of each AI-generated image detector evaluated on five different generators.}
    \label{fig:motivation}
\end{figure}

To address the malicious use of AI-generated images, developing detectors is an effective technique. Current research improves the performance of image detection by enhancing artifact feature extraction algorithms and optimizing classification networks \cite{tan2024frequency,tan2024rethinking,yan2024sanity}. However, the rapid evolution of generative models such as StableDiffusion \cite{rombach_2022_cvpr} and FLUX \cite{flux2024} has introduced greater diversity in image generator styles and distribution shifts. This causes existing detectors to suffer significant performance degradation and eventual failure when confronted with unseen models. As shown in Figure~\ref{fig:motivation}, traditional detectors struggle to adapt to evolving AI-generated image distributions, posing a key bottleneck for real-world applications.

Based on the above issue, adapting detectors to newly emerging image types has become a crucial approach. Traditional fine-tuning methods \cite{cui2018large,kolesnikov2020big,shi2023unified} typically rely on large-scale datasets generated by evolving generative models to preserve detection performance across varying distributions. Meanwhile, continual learning methods \cite{kirkpatrick2017overcoming,kim2021cored,azizpour2024e3} utilize the techniques like regularization, knowledge distillation to maintain the detection performance on the older generative models from another perspective.

However, both categories of methods still face two major challenges that remain insufficiently addressed: \textbf{C1: Low Efficiency.} The efficiency of updating AI-generated image detectors is mainly limited by training data volume and framework complexity. On the data side, achieving sample diversity typically requires frequent retraining and inference with newly generative models, leading to significant computational and storage overhead. On the framework side, updates typically involve structural modifications or multi-stage training to adapt to evolving image distributions, complicating implementation and reducing scalability. These challenges collectively hinder timely and flexible detector updates, especially as generative models evolve rapidly. \textbf{C2: Catastrophic Forgetting.} AI-generated image detector updating methods often prioritize new task data while neglecting retention of prior knowledge, causing models to lose detection ability on previous distributions when adapting to new image features. This reduces accuracy and generalization on earlier generative models. As generative models evolve, relying solely on new data exacerbates knowledge loss, and constraints on storing historical data due to privacy and storage further worsen catastrophic forgetting.



To tackle the above challenges, we propose LiteUpdate, a lightweight framework for updating AI-generated image detectors under iteratively evolving generators. LiteUpdate first adopts a representative sample selection module, leveraging image confidence and loss gradients to identify discriminative features and precisely select samples near the decision boundary. This approach improves learning efficiency and detection accuracy on new distributions using a limited number of AI-generated images, effectively addressing the low efficiency of detector updates. LiteUpdate also integrates a model merging module that fuses weights from multiple fine-tuning trajectories, including pre-trained, representative, and random updates. This approach preserves adaptability to new generators while mitigating catastrophic forgetting of prior knowledge. By integrating two modules, LiteUpdate efficiently balances adaptability to new generators and prior knowledge retention in a lightweight manner.

We conduct extensive experiments on real-world and generated images from diverse sources to evaluate the effectiveness of LiteUpdate. Each detector is firstly pre-trained till convergence to ensure a fair comparison on StableDiffusion 2.1. Experimental results demonstrate that LiteUpdate consistently enhances the performance of different detectors. In particular, for AIDE, the average detection accuracy on Midjourney \cite{midjourney2024} increased from 87.63\% to 93.03\%, representing a relative gain of 6.16\%.

In summary, our work has the following contributions:
\begin{itemize}
\item We propose LiteUpdate, a lightweight adaptive framework for updating AI-generated image detectors with the rapid evolution of image generative models.  
\item We propose a representative sample selection module that efficiently identifies key samples to enhance adaptability to novel distributions. 
\item We employ a model merging module to adapt to new generators while effectively preventing catastrophic forgetting of prior knowledge. 
\item This achieves unprecedented universality for detection frameworks, successfully demonstrating compatibility with various detector such as AIDE, NPR and FreqNet.  

\end{itemize}
\begin{figure*}[t]
    \centering
    \includegraphics[width=\textwidth, height=0.58\textwidth]{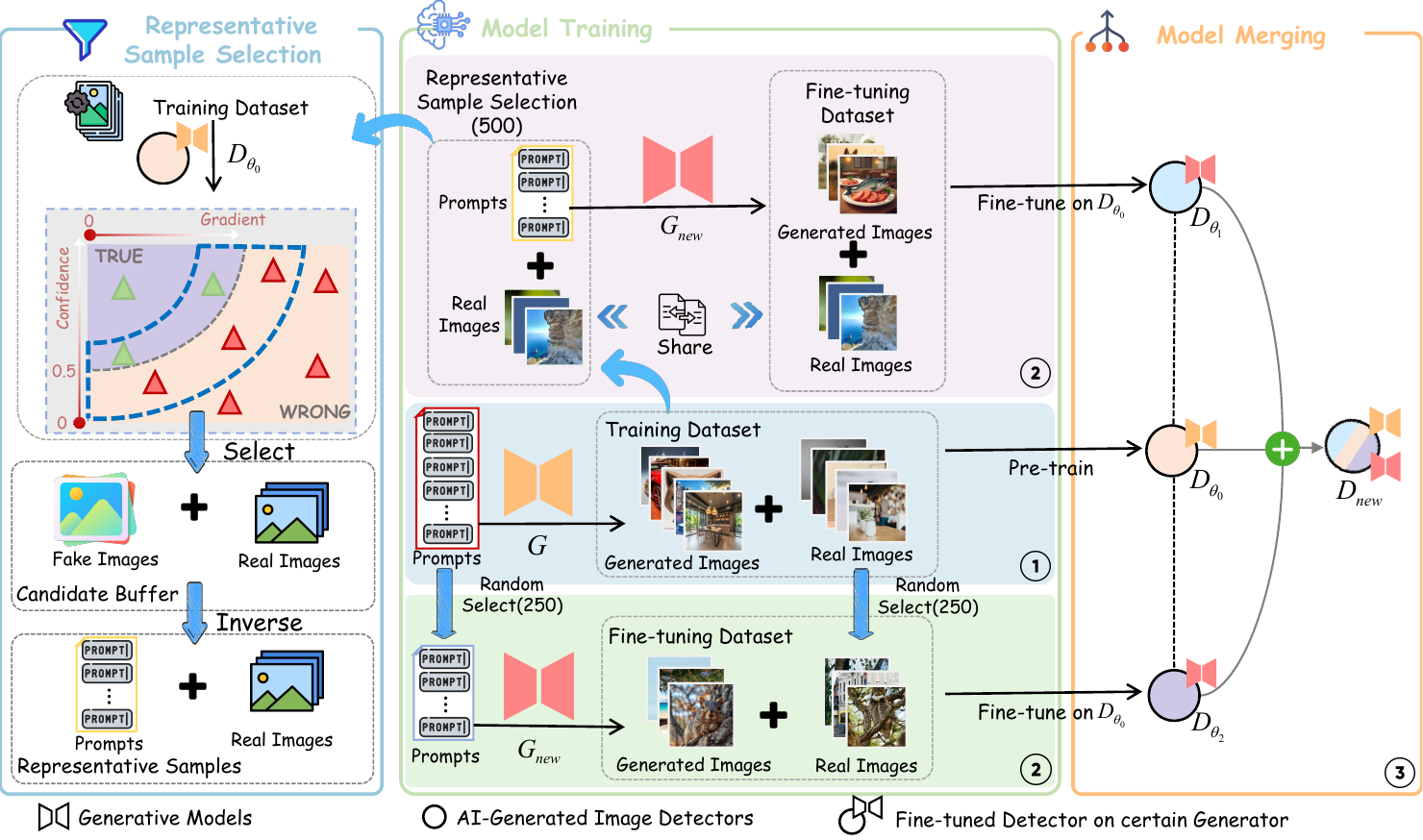}
    \caption{The overview of LiteUpdate. It consists the representative sample selection module and the model merge module. In each step, $D_{\theta_0}$ is the pre-trained detector on generator $G$, while $D_{\theta_1}$ and $D_{\theta_2}$ is the fine-tuned detector on the genrator $G_{new}$ with the random and representative selection module. $D_{new}$ is the updated detector with the LiteUpdate framework.}
    \label{fig:pipeline}
\end{figure*}

\section{Related work}

\subsection{Image Generative Models}
Recent years have witnessed a proliferation of innovative generative models, marking a period of exceptionally rapid development. Image generative models seek to produce high-fidelity images from natural language prompts. To enhance visual quality and semantic consistency, numerous approaches have been proposed in recent years. Early generative models, such as Text-conditional GAN \cite{reed2016generative}, StackGAN \cite{zhang2017stackgan}, and MirrorGAN \cite{qiao2019mirrorgan}, employed GAN-based architectures with attention and multi-stage refinement. With the advent of diffusion models, leading approaches such as StableDiffusion \cite{rombach2022high}, Midjourney \cite{midjourney2024}, DALL-E 3 \cite{ramesh2022hierarchical}, and FLUX \cite{flux2024} excel in photorealism and compositional accuracy. These advance generators present new challenges for updating AI-generated image detectors.

\subsection{AI-Generated Image Detectors}
The challenge of detecting AI-generated images has garnered significant attention over the years. Initial research predominantly concentrated on spatial domain artifacts-leveraging cues \cite{o2012exposing,mccloskey2018detecting,mccloskey2019detecting}. To overcome the poor generalization, learning-based detectors were introduced \cite{wang2020cnn,frank2020leveraging,tan2024frequency}. FreqNet \cite{tan2024frequency} forces the detector to focus on high-frequency representations via High-Frequency Representation and a Frequency Convolutional Layer module. Meanwhile, some efforts have pursued generalization through innovative and often multimodal techniques. NPR \cite{tan2024rethinking} proposes Neighboring Pixel Relationships (NPR), designed to capture local up-sampling artifacts from image pixels. AIDE \cite{yan2024sanity} presents a simple mixture-of-expert model to discern AI-generated images based on both low-level pixel statistics and high-level semantics.

\subsection{AI-Generated Image Detector Update Techniques}
Given different artifacts left by various newly generators, it is crucial for image detectors to continually update their knowledge. Traditional fine-tuning techniques \cite{cui2018large,kolesnikov2020big,shi2023unified} often require extensive datasets produced by evolving generators to ensure consistent detection performance across different distribution shifts. In contrast, continual learning methods \cite{kirkpatrick2017overcoming,kim2021cored,azizpour2024e3} adopt the techniques like regularization, knowledge replay and distillation to preserve the detection performance on the older generative models. E3 \cite{azizpour2024e3} constructs some expert embedders through transfer learning to capture different artifacts from generators and then use them to train an Expert Knowledge Fusion Network (EKFN). However, these existing updating techniques are still facing the problems of low efficiency and catastrophic forgetting. Consequently, by integrating representative sample selection and model merging modules, the LiteUpdate framework efficiently balances adaptation to new generators with prior knowledge retention in a lightweight manner.
\section{Method}
In this section, we define the task of updating AI-generated image detectors and outline the two key modules of the LiteUpdate framework, as shown in Figure~\ref{fig:pipeline}. 

\subsection{Problem Formulation and Overview}
The goal of updating AI-generated image detectors is to a adapt pre-trained model to emerging generative models and evolving application scenarios. Formally, given a detector $D_{\theta_0}$ pre-trained on a mixed dataset of real-world images and AI-generated images produced by the generative model $G$, $D_{\theta_0}$ achieves the goal to detect images generated by $G$. As new generative model $G_{new}$ emerges, the distribution of their outputs may differ significantly from the original training data, causing $D_{\theta_0}$ to degrade in performance. To remain the effectiveness, the detector must be continuously updated. However, retraining from scratch is computationally expensive and risks erasing prior knowledge. Therefore, we aim to update $D_{\theta_0}$ using only a limited number of samples from $G_{new}$, ensuring the updated detector $D_{new}$ can accurately detect images from both $G$ and $G_{new}$.

In the LiteUpdate framework, it consists of representative sample selection module and the model merging module. Specifically, we leverage the pre-trained detector $D_{\theta_0}$ to select representative samples $\mathcal{I}_R$ at the decision boundary, using confidence and gradient information. Then, by integrating the weights of the pre-trained detector $D_{\theta_0}$, the detector $D_{\theta_1}$ fine-tuned on $\mathcal{I}_R$, and the detector $D_{\theta_2}$ fine-tuned on random samples, we obtain the updated detector $D_{new}$.

\subsection{Representative Sample Selection}
In the context of continuously evolving generative models, traditional fine-tuning methods typically involve randomly selecting a set of prompts from the training data of the pre-trained detector $D_{\theta_0}$ and inputting them into the new generative model to produce AI-generated images. Then, it simultaneously samples the real images together to ensure broader distribution coverage. To efficiently update the detector in scenarios involving continuously evolving generators, we first propose a representative sample selection module based on the confidence values of AI-generated images and the gradients of the loss function. By leveraging these two indicators to capture discriminative features, LiteUpdate effectively identifies critical samples near the decision boundary, providing the detector with the most informative and suitable for fine-tuning samples.


\subsubsection{\textbf{Confidence-based Uncertainty Estimation.}} 
For each input image $x_i$, the pre-trained detector $D_{\theta_0}$ extracts features and outputs a probability distribution over the two labels $j \in \{0, 1\}$ via a classification head, enabling image discrimination. The formal definition is as follows:
\begin{equation}
D_{\theta_0}(x_i) = \left[ P(y = 0 \mid x_i; \theta),\; P(y = 1 \mid x_i; \theta) \right]
\end{equation}

Then, for each $x_i$, its confidence value $c_i$ is defined as the predicted probability assigned by the pre-trained detector $D_{\theta_0}$ to its ground-truth label $y_i$:
\begin{equation}
c_i = D_{\theta_0}(x_i)[y_i]
\end{equation}

The confidence value provides an intuitive measure of the certainty of pre-trained detector in image classification and, to some extent, reflects the informational complexity of learned samples. High-confidence samples are typically easy for the detector to recognize and have been well learned, whereas lower-confidence samples tend to exhibit greater informational content and structural complexity, making them more challenging for the detector to learn. Notably, samples with extremely low confidence are often difficult for the detector to learn and may even represent data noise. Excessive attention to such outliers can easily lead to overfitting or bias. Therefore, we focus on representative samples near the decision boundary.


Based on this approach, we rank all training images in order of confidence value. Given that AI-generated image detection is a binary classification task with a decision boundary of 0.5, the most informative boundary samples fall into two types: those with confidence below 0.5 are negatives, while those above 0.5 are positives. Considering that the decision boundary of detector may fluctuate, it is important to maintain a balanced selection of both positive and negative samples. Therefore, we select positive and negative boundary samples in a certain proportion to construct the candidate buffer $\mathcal{I}$ with images, formally defined as follows:
\begin{equation} \label{eq:confidence}
\mathcal{I} = \left\{ 
    x_i\,\middle|\,
    \begin{aligned}
    &\left(\mathrm{Rank}_{\downarrow}(c_i \mid c_i < 0.5)
 \leq \lceil k_n N \rceil \right) \\
    &\vee\ \left(\mathrm{Rank}_{\uparrow}(c_i \mid c_i \geq 0.5)
\leq \lceil k_p N \rceil \right)
    \end{aligned}
\right\}
\end{equation}
where $N$ denotes the representative sample size specified in our paper, which is set to 500. Specifically, $\mathrm{Rank}_{\uparrow}(c_i)$ refers to the index of $c_i$ in the confidence scores sorted in ascending order, while $\mathrm{Rank}_{\downarrow}(c_i)$ denotes the reverse index. The parameters $k_p$ and $k_n$ represent the proportions of positive and negative boundary samples, respectively, and satisfy $k_p + k_n = 1$. By leveraging confidence values to mine discriminative features, it becomes possible to more effectively identify key images with higher informational content.







\subsubsection{\textbf{Gradient-based Sensitivity Analysis.}} Relying solely on confidence values is insufficient to effectively identify the most critical samples for updating detector. To enhance the precision of sample selection, LiteUpdate introduces a gradient-based sensitivity metric $g_i$, which aims to identify high-value samples for fine-tuning from the pool of positive and negative boundary samples. Specifically, for each image $x_i$, we compute its gradient sensitivity score $g_i$, which is defined as the $L_2$ norm of the gradient of the loss function $\mathcal{L}(D_{\theta_0}(x_i), y_i)$ with respect to all trainable parameters:
\begin{equation} 
g_i = \left\| \nabla_\theta \mathcal{L}(D_{\theta_0}(x_i), y_i) \right\|_2
= \sqrt{ \sum_j \left( \frac{\partial \mathcal{L}}{\partial \theta_j} \right)^2 }
\end{equation}
where $L$ denotes the loss function specific to each AI-generated image detector. Gradient sensitivity quantifies the extent to which an image influences changes in the detector loss within the parameter space. A higher value indicates that the image is more sensitive to small perturbations in the detector parameters, thereby facilitating more effective parameter updates toward the optimization objective and improving the efficiency of detector training. 

Building on the above approach, we further update the images in the candidate buffer $\mathcal{I}$ described in Equation~\eqref{eq:confidence}. For samples with $c_i \geq 0.5$, we use gradient sensitivity as the selection criterion and rank the images in the order of their gradient values, as defined below:
\begin{equation}
\mathcal{I} = \left\{ 
    x_i\,\middle|\,
    \begin{aligned}
    &\left(\mathrm{Rank}_{\downarrow}(c_i \mid c_i < 0.5)
 \leq \lceil k_n N \rceil \right) \\
    &\vee\ \left(\mathrm{Rank}_{\downarrow}(g_i \mid c_i \geq 0.5)
\leq \lceil k_p N \rceil \right)
    \end{aligned}
\right\}
\end{equation}
where $\mathrm{Rank}_{\uparrow}(g_i)$ refers to the index of $g_i$ in the gradient sensitivity score sorted in ascending order, while $\mathrm{Rank}_{\downarrow}(g_i)$ denotes the reverse index.

In addition, negative boundary samples are misclassified cases, and selecting them based on gradient sensitivity may drive the parameters of detector in the wrong direction. Therefore, we do not use gradient sensitivity as a selection criterion for these samples.



\subsubsection{\textbf{Representative Sample.}} After obtaining the candidate buffer image set $\mathcal{I}$, we divide it into subsets of real-world $\mathcal{I}_{real} $and AI-generated images $\mathcal{I}_{AI}$:
\begin{equation}
\mathcal{I} = \mathcal{I}_{real} \cup \mathcal{I}_{AI}
\end{equation}
Then, we extract the corresponding prompts from $\mathcal{I}_{AI}$ and use the new generator $G_{new}$ to produce new images, forming the set $\mathcal{I}_{new}$. Finally, with our representative sample selection module, we obtain the representative sample set $\mathcal{I}_R$:
\begin{equation}
\mathcal{I}_{R} = \mathcal{I}_{real} \cup \mathcal{I}_{new}
\end{equation}


\subsection{Model Merging}

To preserve the overall detection performance across both old and new generators, we integrate the complementary knowledge learned by the three detectors introduced earlier: the pre-trained detector, the representative-sample fine-tuned detector, and the random-sample fine-tuned detector. 

Inspired by model merging techniques~\cite{wortsman2022model}, we implement this integration by linearly combining their parameter weights. Specifically, let $\theta_0$ denote the pre-trained weights, $\theta_1$ the representative-sample fine-tuned weights, and $\theta_2$ the random-sample fine-tuned weights. The merged weights are defined as:
\begin{align}
\theta_{\text{merged}} = (1 - 2k)\cdot \theta_0 + k \cdot \theta_1 + k \cdot \theta_2
\label{eq:model_merge}
\end{align}
where $k \in [0, 0.5]$ is a balancing coefficient that adjusts the influence of the two fine-tuned weights relative to the pre-trained weights. This symmetric interpolation form ensures balanced integration of the two adaptation strategies, while anchoring the merged weights to $\theta_0$ helps prevent catastrophic forgetting and overfitting to the new generator.

\begin{figure*}[t]
    \centering
\includegraphics[width=0.78\textwidth, height=0.34\textwidth]{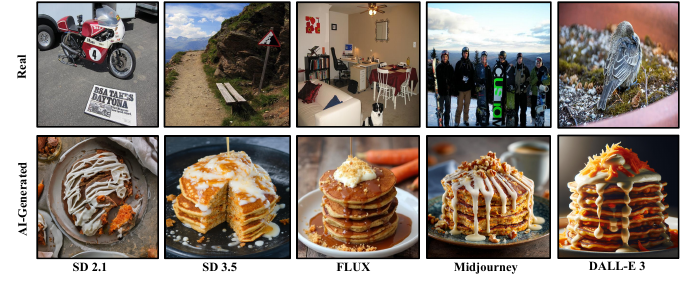}
    \caption{Examples of real-world and AI-generated images from our dataset. AI-generated images are from generators such as SD 2.1, SD 3.5, FLUX, Midjourney, and DALL-E 3 with the prompt ``Carrot Cake Pancakes topped with Maple Cream Cheese drizzle". We can see that all samples exhibit similar quality in terms of content preservation.}
    \label{fig:substitution}
\end{figure*}
\section{Experiments}
In this section, we conduct experiments to comprehensively evaluate the challenges in adapting AI-generated images detectors. Some effectiveness evaluation are performed to validate the performance gains enabled by LiteUpdate. Furthermore, ablation studies are carried out to assess the contribution of each module within LiteUpdate framework.
\subsection{Experiments Setup}
\subsubsection{\textbf{Environment.}} Experiments are conducted on an Ubuntu 22.04 server with an Intel Xeon Gold 6330 CPU, 500GB RAM, 8TB storage, and 8 NVIDIA RTX A6000 GPUs.
\subsubsection{\textbf{Dataset Construction.}} We compile a dataset from existing resources in order to train and evaluate the AI-generated image detector with different strategies. Real-World images are selected from the MSCOCO dataset \cite{lin2014microsoft}, renowned for its high visual quality and rich variety of contents. AI-generated images are produced by Stable Diffusion 2.1, StableDiffusion 3.5 \cite{rombach_2022_cvpr}, FLUX \cite{flux2024}, Midjourney \cite{midjourney2024}, and DALL-E 3 \cite{ramesh2022hierarchical} using text prompts from the LAION-AESTHETICS dataset \cite{laion-aesthetics}. All images are generated at a square resolution of $512 \times 512$ using default settings. Figure~\ref{fig:substitution} shows examples of real-world and AI-generated images in our dataset. For pre-trained detector training, we collected a balanced set of 64K, 8K, and 8K images for training, validation, and testing, respectively.

\subsubsection{\textbf{Selected Detectors and Baselines.}} We select several recently proposed and high-performing generative image detectors as our backbone models to evaluate the effectiveness of LiteUpdate including FreqNet \cite{tan2024frequency}, NPR \cite{tan2024rethinking}, and AIDE \cite{yan2024sanity}. In addition, we adapt both the state-of-the-art continual learning method E3 \cite{azizpour2024e3} and detectors fine-tuned on 1,000 randomly selected samples as our baselines.

\subsubsection{\textbf{Implementation Details and Metrics.}} In real-world scenarios, detectors for various generative models are typically pre-trained on specific datasets. To simulate this scenario, we pre-train each detector using AI-generated images from StableDiffusion 2.1 and real-world images from the MSCOCO dataset. For brevity, we refer to StableDiffusion as ``SD" in subsequent experiments and analyses.

In the LiteUpdate framework, the representative sample selection module uses a 0.1:0.9 ratio of positive and negative boundary samples, and the model merging module employs a balancing coefficient $k=0.2$ for updating. Each detector is first pre-trained for 10 epochs with a batch size of 32, and then fine-tuned for 5 epochs on different generative models using 500 images with a batch size of 8. The optimizer, learning rate, and other hyperparameters are set to the default values for each detector.

We use the data augmentations including random Gaussian blur ($\sigma\sim\mathrm{Uniform}(0.1, 3.0)$) and random JPEG compression ($QF\sim\mathrm{Uniform}(30, 100)$) to improve the robustness of each detector during the training procedure. Each augmentation is conducted with 10\% probability. During the test phase, we adopt the JPEG compression ($QF = 50$) and random Gaussian blur ($\sigma = 1.0 $) with 50\% probability to better align with real-world scenarios. 

Additionally, we use the detection accuracy to evaluate the performance of both the proposed and baseline methods.

\setlength{\tabcolsep}{1mm}  
\begin{table*}[htbp]
  \centering
  {\fontsize{9pt}{11pt}\selectfont
  \begin{tabular}{lccc ccc ccc ccc}
      \toprule
      \multirow{2}{*}{Method} 
      & \multicolumn{3}{c}{SD 3.5} 
      & \multicolumn{3}{c}{FLUX} 
      & \multicolumn{3}{c}{Midjourney} 
      & \multicolumn{3}{c}{DALL-E 3} \\
      \cmidrule(lr){2-4} \cmidrule(lr){5-7} \cmidrule(lr){8-10} \cmidrule(lr){11-13}
      & SD 2.1 & SD 3.5 & Avg  
      & SD 2.1 & FLUX & Avg 
      & SD 2.1 & Midjourney & Avg  
      & SD 2.1 & DALL-E 3 & Avg   \\
      \midrule
      E3              & 80.94 & 63.26 & 72.10  & 79.74 & 63.41 & 71.58  & 72.68 & 82.60 & 77.64  & 78.50 & 79.12 & 78.81 \\
      \midrule
      Pre-train-AIDE            & 98.31 & 78.90 & 88.61 & 98.31 & 76.76 & 87.54 & 98.31 & 77.69 & 88.00 & 98.31 & 84.54 & 91.43 \\
      Random-AIDE            & 89.70 & 88.12 & 88.91  & 86.80 & 86.05 & 86.43  & 87.96 & 87.30 & 87.63  & 97.14 & 90.12 & 93.63 \\
      \rowcolor[gray]{0.9}
      \textbf{LiteUpdate-AIDE} & 96.25 & 89.74 & \textbf{92.99}  & 93.63 & 89.59 & \textbf{91.61}  & 95.69 & 90.36 & \textbf{93.03}  & 97.44 & 92.91 & \textbf{95.18} \\
      \midrule
      Pre-train-FreqNet         & 86.76 & 65.91 & 76.34 & 86.76 & 60.85 & 73.81  & 86.76 & 58.61 & 72.69 & 86.76 & 62.41 & 74.59 \\
      Random-FreqNet         & 79.28 & 71.20 & 75.24  & 70.14 & 65.01 & 67.58  & 82.44 & 70.67 & \textbf{76.55}  & 73.21 & 72.18 & 72.70 \\
      \rowcolor[gray]{0.9}
      \textbf{LiteUpdate-FreqNet} & 86.08 & 67.03 & \textbf{76.55} & 84.71 & 67.06 & \textbf{75.89} & 85.61 & 66.29 & 75.95 & 86.26 & 63.13 & \textbf{74.70} \\
      \midrule
      Pre-train-NPR             & 91.35 & 62.10 & 76.73 & 91.35 & 55.14 & 73.25 & 91.35 & 57.87 & 74.61 & 91.35 & 57.56 & 74.46 \\
      Random-NPR             & 85.84 & 77.73 & \textbf{81.78}  & 69.50 & 66.98 & 68.24  & 80.33 & 76.17 & 78.25  & 71.23 & 81.39 & 76.31 \\
      \rowcolor[gray]{0.9}
      \textbf{LiteUpdate-NPR}  & 89.49 & 69.99 & 79.74  & 74.21 & 67.80 & \textbf{71.01}  & 87.33 & 69.67 & \textbf{78.50}  & 87.68 & 75.20 & \textbf{81.44} \\
      \bottomrule
  \end{tabular}
    \caption{The detection accuracy (\%) of each AI-generated image detector. Pre-train refers to the different detector trained on the generator $G$ (SD 2.1 in this study)  without any updates. Random denotes the detector fine-tuned with randomly sampled images while LiteUpdate represents our lightweight framework.}
  \label{tab:sd_five_block_eval}
  }
\end{table*}

\begin{figure}[t]
    \centering
    \includegraphics[width=0.42\textwidth, height=0.28\textwidth]{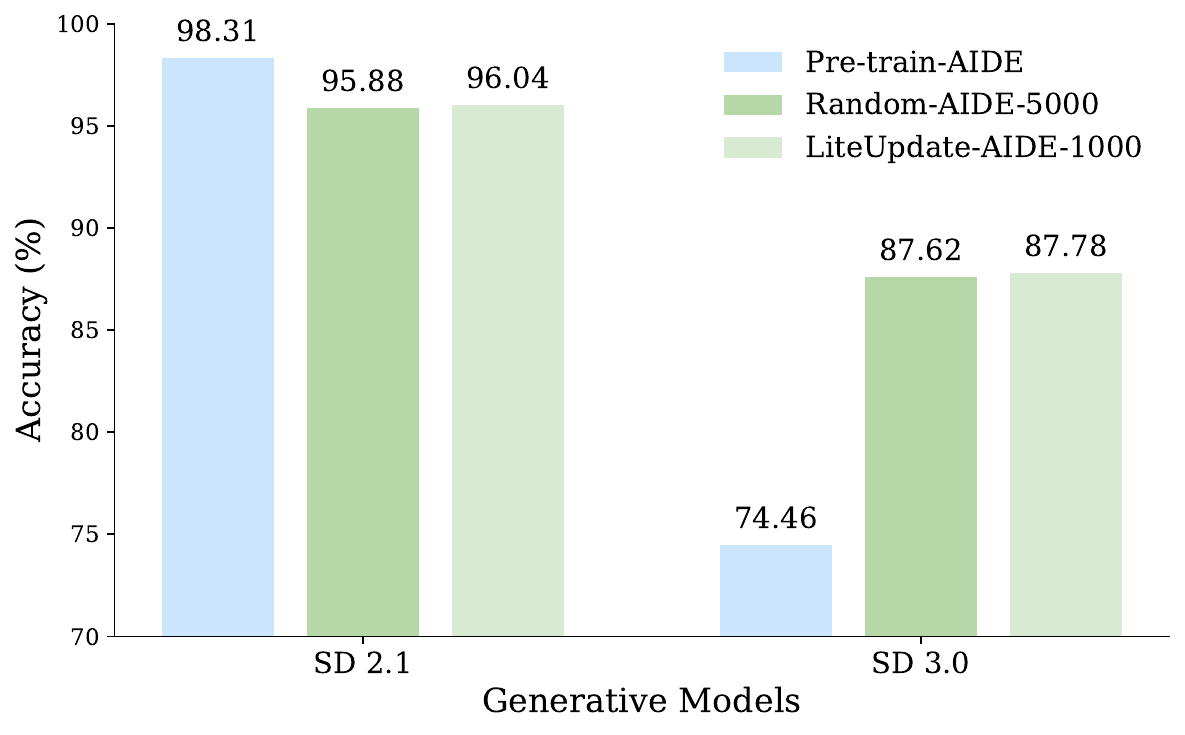}
    \caption{Efficiency comparison between random samples fine-tuning and the LiteUpdate on AIDE detector while the newly generator is SD 3.0. The LiteUpdate achieves superior detection performance using only 1,000 samples, surpassing the detector fine-tuned with random 5,000 samples.}
    \label{fig:Thereshold}
\end{figure}
\subsection{Main Results} 



To comprehensively evaluate LiteUpdate, we conduct a series of experiments on detection accuracy, data efficiency, and multi-generator adaptability.

\textbf{Effectiveness of Liteupdate.} 
The detection performance of various methods on different generative models is shown in Table~\ref{tab:sd_five_block_eval}. After applying the random sample fine-tuning strategy, AI-generated image detectors exhibit improved performance on new generative models compared to the pre-trained detectors, but suffer a substantial decline in performance on the older generative model (SD 2.1). For example, after random sample fine-tuning, the detection accuracy of the NPR detector on DALL-E 3 increased to 81.39\%, representing an improvement of approximately 24\%. However, its accuracy on SD 2.1 dropped to 71.23\%, a decrease of 20.12\%, resulting in only a modest overall improvement of 1.85\%. In addition, most detectors exhibited an overall performance decline after random sample fine-tuning; for instance, the overall performance of FreqNet on FLUX decreased by 6.23\%. The results indicate that fine-tuning detectors only with new AI-generated images is insufficient for effective updates as generative models evolve.

Compared with random sample fine-tuning, LiteUpdate effectively mitigates the problem of forgetting previous knowledge. For example, when applied to AIDE, LiteUpdate improves accuracy on Midjourney by 12.67\% over the pre-trained detector, with only a 2.62\% decrease on SD2.1, resulting in an average accuracy gain of 5.03\%. These results demonstrate that LiteUpdate can adapt to new generators while retaining prior knowledge.

Compared with the state-of-the-art continual learning method E3, LiteUpdate achieves higher average detection accuracy on most generative models, demonstrating superior performance with a more lightweight framework.




\textbf{Efficiency of Liteupdate.} Meanwhile, to demonstrate the superiority of LiteUpdate, we conduct a fine-tuning threshold experiment on AIDE detector while newly generator is SD 3.0. As shown in Figure~\ref{fig:Thereshold}, the AIDE detector achieves an average detection accuracy of 91.75\% when fine-tuned with 5,000 randomly selected samples, which is comparable to the performance achieved by the LiteUpdate framework using only 1,000 representative samples. This clearly demonstrates the efficiency of LiteUpdate in achieving high detection performance with significantly fewer samples.

\begin{figure}[t]
\centering
\includegraphics[width=0.47\textwidth, height=0.32\textwidth]{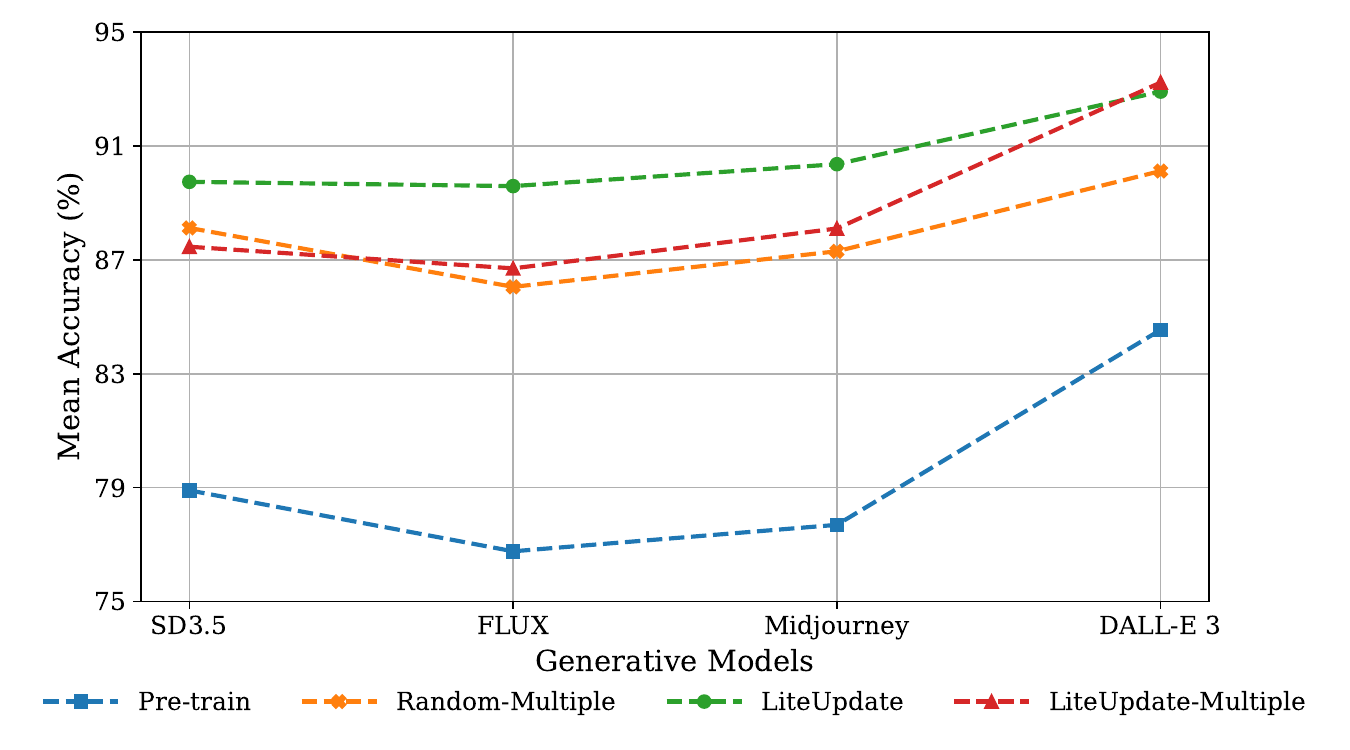}
\caption{The mean detection accuracy of different strategies on AIDE across both old and newly generative models under a multi-generator setting. LiteUpdate refers to our framework designed to adapt to a single newly generator, while LiteUpdate-Multiple extends this capability to simultaneously adapt to multiple emerging generators.}
\label{fig:expand}
\end{figure}

\textbf{Adapting to Multiple Evolving Generators.} In this experiment, we try to investigate the ability of LiteUpdate to adapt to multiple evolving generators in the real world scenario. After selecting representative samples, their prompts are evenly distributed across different generators and combined with representative real-world images to fine-tune the pre-trained AIDE detector. As shown in Figure~\ref{fig:expand}, LiteUpdate achieves consistently high detection performance across all generators, significantly outperforming the random sample selection strategy. Notably, LiteUpdate-Multiple achieves a detection accuracy of 93.23\% on DALL-E 3, outperforming LiteUpdate in the single-generator setting. This demonstrates that LiteUpdate not only enables lightweight updates in single-generator scenarios but also generalizes well to multi-generator environments.


\subsection{Ablation Studies}
To evaluate the effectiveness of each module in the LiteUpdate framework, we conduct extensive ablation studies using the AIDE detector while the newly generator is SD 3.5.


\textbf{Ablation for Negative Boundary Sample Ratio.} To evaluate the effectiveness of the positive and negative boundary sample ratio in the representative sample selection module, we fix the total number of representative samples at 500 and vary the negative boundary sample ratio from 0.1 to 1.0. The results are shown in Table~\ref{tab:Sample_ratio_ablation}. The results show that the best performance is achieved when the negative boundary sample ratio is 0.9, yielding the highest detection accuracy of 94.40\% and 91.86\%, respectively, with an average of 93.13\%. This indicates that selecting representative samples with a certain ratio of positive and negative boundary samples enhances the error correction and maintain a smooth transition of the decision boundary. It can avoid overfitting and improve the overall detection performance. However, the result also shows that an excessive number of positive boundary samples will lead to the poor detection performance. We consider that these samples may introduce redundancy and reduce sensitivity to borderline cases. Finally, we recommend the negative boundary sample ratio $k_n = 0.9$ during the representative sample selection module.


\begin{table}[t]
\centering
\setlength{\tabcolsep}{10pt}
{\fontsize{9pt}{11pt}\selectfont
\begin{tabular}{ccccc}
    \toprule
    \textbf{Ratio}($k_n$) & \textbf{SD 2.1} & \textbf{SD 3.5} & \textbf{Avg} \\
    \midrule
    0.1 & 86.36 & 85.46 & 85.91 \\
    0.3 & 91.31 & 88.89 & 90.10 \\
    0.5 & 91.96 & 90.05 & 90.85 \\
    0.7 & 91.58 & 90.38 & 90.98 \\
    0.9 & \textbf{94.40} & \textbf{91.86} & \textbf{93.13} \\
    1.0 & 90.80 & 89.73 & 90.26 \\
    \bottomrule
\end{tabular}
\caption{The results under different negative boundary sample ratios $k_n$ during representative sample selection module.}
\label{tab:Sample_ratio_ablation}
}
\end{table} 

\textbf{Ablation for Merging block and coefficient.} We conduct the balancing coefficient $k$ on the model merging module selection experiment in the equation~\eqref{eq:model_merge}. As shown in the Table~\ref{tab:error_ratio_ablation_reduced}, we vary $k$ from 0.0 to 0.5 and evaluate the detection accuracy on SD 2.1 and SD 3.5. When $k = 0.2$, the merged detector achieves the highest average accuracy of 93.30\%, with detection accuracy reaching 96.85\% on SD 2.1 and 89.74\% on SD 3.5. Compared with $k=0.0$, which yields an average accuracy of 88.61\%, the performance improves significantly. As $k$ increases further beyond 0.2, we observe a gradual decrease in average accuracy. Therefore, $k=0.2$ provides the most balanced and effective merging module between retaining base knowledge and incorporating new generator-specific adaptations.

After selecting the balancing coefficient $k$, we use $k=0.2$ to conduct the ablation experiment of model merging block. As shown in Table~\ref{tab:modules}, removing either the representative sample fine-tuned detector $D_{\theta_1}$ or the random sample fine-tuned detector $D_{\theta_2}$ results in obvious performance degradation in terms of the mean accuracy (87.22\% and 85.77\% vs 92.99\%). The accuracy on the old (SD 2.1) or new (SD 3.5) generative model is obviously lower than the merging detector $D_{new}$. The degree of forgetting is even lower than $D_{new}$ for $D_{\theta_1}$ or $D_{\theta_2}$ by approximately 10\%. All in all, merging both these two parts effectively balance precise adaptation with distributional robustness.

\begin{table}[t]
\centering
\setlength{\tabcolsep}{10pt}
{\fontsize{9pt}{11pt}\selectfont
\begin{tabular}{cccc}
    \toprule
    \textbf{Coefficient($k$)} 
    & \textbf{SD 2.1} & \textbf{SD 3.5} & \textbf{Avg} \\
    \midrule
    0.0 & 98.31 & 78.90 & 88.61 \\ 
    0.1 & 97.63 & 85.04 & 91.33 \\
    0.2 & 96.85 & 89.74 & \textbf{93.30} \\
    0.3 & 93.51 & 89.61 & 91.76 \\
    0.4 & 92.45 & 89.44 & 90.95 \\
    0.5 & 89.58 & 88.76 & 89.17 \\ 
    \bottomrule
\end{tabular}
\caption{Ablation study results for balancing coefficient $k$ in the model merging module.}
\label{tab:error_ratio_ablation_reduced}
}
\end{table}

\setlength{\tabcolsep}{1.6mm}  
\begin{table}[t]
    \centering
    {\fontsize{9pt}{12pt}\selectfont
    \begin{tabular}{ccc|ccc}
        \toprule
        \multicolumn{3}{c|}{\textbf{Component}} & \multicolumn{3}{c}{\textbf{Accuracy}} \\
        \cmidrule(lr){1-3} \cmidrule(lr){4-6}
        \textbf{$D_{\theta_0}$} & \textbf{$D_{\theta_1}$} & \textbf{$D_{\theta_2}$} & \textbf{SD 2.1} & \textbf{SD 3.5} & \textbf{Avg} \\
        \midrule
        \checkmark & \xmark & \xmark & 98.31 & 78.90 & 88.61 \\
        \xmark & \xmark & \checkmark & 87.88 & 86.55 & 87.22 \\
        \xmark & \checkmark & \xmark & 86.50 & 85.03 & 85.77 \\
        \midrule
        \checkmark & \checkmark & \checkmark & 96.25 & 89.74 & \textbf{92.99} \\
        \bottomrule
    \end{tabular}
    \caption{Ablation study results on the model merging between $D_{\theta_0}$, $D_{\theta_1}$ and $D_{\theta_2}$. The balancing coefficient $k = 0.2$ and the negative boundary sample ratio $k_n = 0.9$.}
    \label{tab:modules}
    }
\end{table}

\section{Conclusion}
In this paper, we propose LiteUpdate, a lightweight framework for updating AI-generated image detectors under iteratively evolving generators. By incorporating representative sample selection and model merging modules, LiteUpdate significantly improves detector update efficiency and enhances adaptability to new generators, while effectively mitigating catastrophic forgetting of previous knowledge. Experimental results demonstrate that LiteUpdate consistently enhances the performance of different detectors.

\bibliography{aaai2026}

\end{document}